\documentclass[10pt,journal,compsoc]{IEEEtran}
\ifCLASSOPTIONcompsoc
  \usepackage[nocompress]{cite}
\else
  \usepackage{cite}
\fi

\usepackage{epsfig}
\usepackage{graphicx}
\usepackage{amsmath}
\usepackage{amssymb}

\usepackage{multirow}
\usepackage{stackengine}
\usepackage{xspace}
\usepackage{mathtools,xparse}
\usepackage{breqn}
\usepackage{xcolor}

\usepackage{url}

\newcommand{\norm}[1]{\left\|#1\right\|} 

\makeatletter
\DeclareRobustCommand\onedot{\futurelet\@let@token\@onedot}
\def\@onedot{\ifx\@let@token.\else.\null\fi\xspace}
\def\eg{\emph{e.g}\onedot} 
\def\ie{\emph{i.e}\onedot}

\def\etal{\emph{et al}\onedot}
\makeatother

\begin{document}
%
\title{Revisiting Image-Language Networks for Open-ended Phrase Detection}
\author{Bryan A. Plummer, 
        Kevin J. Shih, 
        Yichen Li, %
        Ke Xu,
        Svetlana Lazebnik,~\IEEEmembership{Senior Member,~IEEE},
        Stan Sclaroff,~\IEEEmembership{Fellow,~IEEE},
        Kate Saenko
\IEEEcompsocitemizethanks{\IEEEcompsocthanksitem Bryan A. Plummer, Yichen Li, Stan Sclaroff, and Kate Saenko are with the Department of Computer Science, Boston University, Boston, MA, 02215. E-mail: \{bplum,liych,sclaroff,saenko\}@bu.edu
\IEEEcompsocthanksitem Kevin J. Shih is with the NVIDIA Corporation, Santa Clara, CA 95051. Email: kshih@nvidia.com
\IEEEcompsocthanksitem Ke Xu and Svetlana Lazebnik are with the Department of Computer Science, University of Illinois at Urbana-Champaign, Urbana, IL 61801. Email: \{kexu6,slazebni\}@illinois.edu}
}

%

\IEEEtitleabstractindextext{%
\begin{abstract}
Most existing work that grounds natural language phrases in images starts with the assumption that the phrase in question is relevant to the image. In this paper we address a more realistic version of the natural language grounding task where we must both identify whether the phrase is relevant to an image \textbf{and} localize the phrase.  This can also be viewed as a generalization of object detection to an open-ended vocabulary, introducing elements of few- and zero-shot detection. We propose an approach for this task that extends Faster R-CNN to relate image regions and phrases.  By carefully initializing the classification layers of our network using canonical correlation analysis (CCA), we encourage a solution that is more discerning when reasoning between similar phrases, resulting in over double the performance compared to a naive adaptation on three popular phrase grounding datasets, Flickr30K Entities, ReferIt Game, and Visual Genome, with test-time phrase vocabulary sizes of 5K, 32K, and 159K, respectively.
\end{abstract}

}

\maketitle

\IEEEdisplaynontitleabstractindextext

%
\IEEEpeerreviewmaketitle

\IEEEraisesectionheading{\section{Introduction}\label{sec:introduction}}


Traditionally, object detection and localization benchmarks have focused on relatively few categories with many samples per each category. As methods tailored for such benchmarks reach maturity, the recognition community is starting to push towards larger-vocabulary, long-tailed scenarios -- see, \eg,~\cite{Gupta19} for a recent example. If we allow object categories to be described by freeform natural language phrases, we arrive at the task of {\em phrase detection}.


Recently there has been a lot of research on various {\em phrase grounding} or {\em localization} scenarios. In the most common scenario, the text query to be localized is assumed to be present in the test image~\cite{ChenICMR2017,ChenICCV2017,fukui16emnlp,hu2015natural,kazemzadeh-EtAl:2014:EMNLP2014,mao2016generation,plummerCITE2017,plummerPLCLC2017,flickrentitiesijcv,rohrbach2015,wangTwoBranch2017,wang2016matching,yehNIPS2017}. The few works that relax this assumption make other simplifications, \eg, limiting the number of negative or distractor images for a query~\cite{Zhang_2017_CVPR}, or limiting queries to the most common phrases in a dataset~\cite{hinamiARXIV2017}. Despite significant gains reported in these and other works, phrase grounding has, so far, failed to yield decisive improvements in downstream applications for which it would seem to be a natural building block, such as text-to-image search, image captioning, or visual question answering. We believe this is due at least in part to the restricted definitions of grounding adopted in existing work.

%

\begin{figure}
\centering
\includegraphics[width=0.43\textwidth]{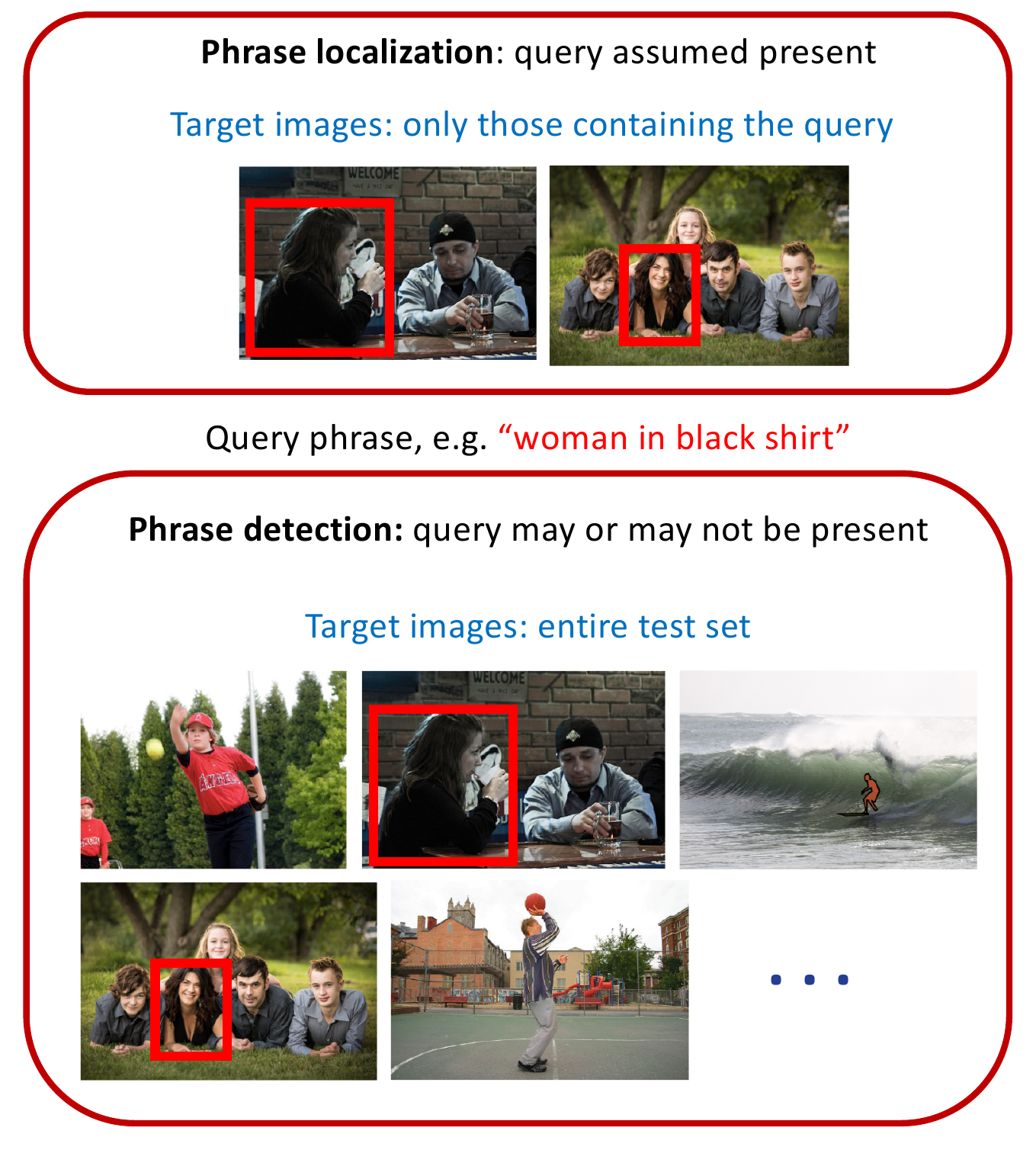}
\caption{In this paper we address the task of phrase detection, where the goal is to identify any image regions related to a query phrase and remove all other candidates.  This is a more challenging version of the phrase localization task addressed in prior work that assumes ground truth image-phrase pairs are given at test time. 
}
\label{fig:motivation}
\end{figure}

In this paper we consider a true {\em phrase detection} benchmark without any simplifications or restrictions.  Given a query phrase, the goal is to identify every image region associated with that phrase within a database of test images. Figure~\ref{fig:motivation} contrasts this definition with the far more popular {\em phrase localization} scenario in which the goal is only to find queries that are present {\em somewhere} in the image. Phrase localization is typically evaluated using accuracy, or the percentage of queries correctly localized. To obtain good performance, it is sufficient to simply identify the relevant region for a given phrase. Crucially, there is no need to ensure consistent calibration of scores across different test images. By contrast, phrase detection is evaluated using average precision, which measures the ability of the model to separate positive regions from negative ones {\em across the entire set of test images} -- a much more demanding criterion.
In other words, a good phrase detector must return region-phrase scores that can be consistently interpreted as probabilities or confidences that the given phrase describes the given region. As our experiments will demonstrate, this is a very hard challenge indeed for most existing methods.



A good phrase detector must also be able to localize phrases well, but a good phrase localizer may be a poor detector.  
This is because in standard phrase localization benchmarks, distinguishing between closely related phrases is usually unnecessary. To give a concrete example, in the Flickr30K Entities benchmark~\cite{flickrentitiesijcv}, 60\% of test images have a single annotated person reference, and 27\% have only two. If we assume every person is annotated in these images and we assigned predicted person boxes to person phrases at random, we would expect to get 79\% of the person references correct.  This is even more pronounced for other phrase types -- \eg, an oracle vehicle detector would get 95\% of vehicle phrases correct since most images only contain a single reference to a vehicle. Thus, phrase localization often degenerates to simply identifying the basic object category of a phrase. Consequently, models trained for phrase localization tend to overfit to these conditions, and have little ability to identify which phrases are actually relevant to an image.  Thus, simple methods such as Canonical Correlation Analysis (CCA)~\cite{hotelling1936relations} tend to perform much better on phrase detection than the state-of-the-art in phrase localization.

To demonstrate this, we benchmark several methods that perform well on the phrase localization task. These include the classical CCA baseline, as well as several state-of-the-art methods including two-branch embedding networks~\cite{wangTwoBranch2017}, the similarity network~\cite{wangTwoBranch2017}, and Query-Adaptive R-CNN~\cite{hinamiARXIV2017}. Although each of these methods outperforms CCA on phrase localization, CCA sometimes \emph{doubles} their performance on the phrase detection task.  

As the analysis of Section \ref{sec:det} will demonstrate, CCA's surprisingly good performance is due to its ability to better discriminate across similar phrases than its neural network counterparts. This makes sense as CCA can be seen as whitening and then aligning the image regions and text features by looking at the entire dataset.  In contrast, the minibatches used to train neural networks only see a tiny portion of the data at a time due to limits of GPU memory. This makes distinguishing between similar phrases difficult, since relatively few are in each minibatch during training.

The results of our experiments inspire us to reconsider CCA's role in vision-language tasks.  Rather than being used only as a baseline, it can be seen as a data whitening or normalization procedure that can be used to initialize the layers of a neural network responsible for mapping together visual and textual representations, instead of using the standard random initialization. 
For a phrase detection model, this means we could fine-tune the CCA weights, which are trained only using positive region-phrase pairs, to make them more discriminative by showing them both positive and negative pairs. This results in a model that does well for both phrase detection and localization. 

In addition to taking a fresh look at the role of CCA for challenging image-language tasks, we evaluate several additional ideas to further boost performance.  First, we use WordNet~\cite{Miller:1995:WLD:219717.219748} to perform positive phrase augmentation (PPA), which identifies valid alternatives to annotated phrases. With this approach, we would consider \emph{a person} to be a positive for a region annotated with \emph{a construction worker} even if they were not annotated as such.  This helps mitigate the annotation sparsity issues in phrase grounding datasets.  Second, to help reduce overfitting to the phrase localization task, we use inverse frequency sampling (IFS), which biases our minibatches to select less common phrases during training.  Many of these rarer phrases refer to fine-grained concepts, and by including them more often we encourage our model to learn how to separate them.

Our contributions are summarized as follows. We argue that unrestricted {\em phrase detection} is a more challenging and meaningful task for visual grounding than {\em phrase localization}, in which the query phrase is already assumed to be present in the image. 
In the proposed detection framework, we perform a comparative evaluation of several recent phrase localization methods. We evaluate on three popular grounding datasets, Flickr30K Entities~\cite{flickrentitiesijcv}, ReferIt Game~\cite{kazemzadeh-EtAl:2014:EMNLP2014}, and Visual Genome~\cite{krishnavisualgenome}, with test set phrase vocabulary sizes of 5K, 32K, and 159K, respectively. We rank methods using mean average precision (mAP) across these entire vocabularies, further broken down by phrase frequencies. Perhaps surprisingly, we find that the relative standing of different approaches on the localization and detection tasks is quite different. In particular, state-of-the-art phrase localization models tend to overfit on localization and do relatively poorly on detection, while seemingly ``simpler'' CCA baselines are better able to discriminate between similar phrases and produce scores that are more predictive of the presence of a phrase in the image. Ultimately, we obtain our best detection performance by fine-tuning a CCA-initialized model, suggesting that CCA may be best thought of as a basic data alignment or normalization procedure that improves performance for cross-modal tasks. Our code is publicly available\footnote{\url{https://github.com/BryanPlummer/phrase_detection}} to encourage further research on phrase detection. 



\begin{figure*}
\centering
\includegraphics[width=\textwidth,trim=0cm 3.8cm 10.8cm 0cm,clip]{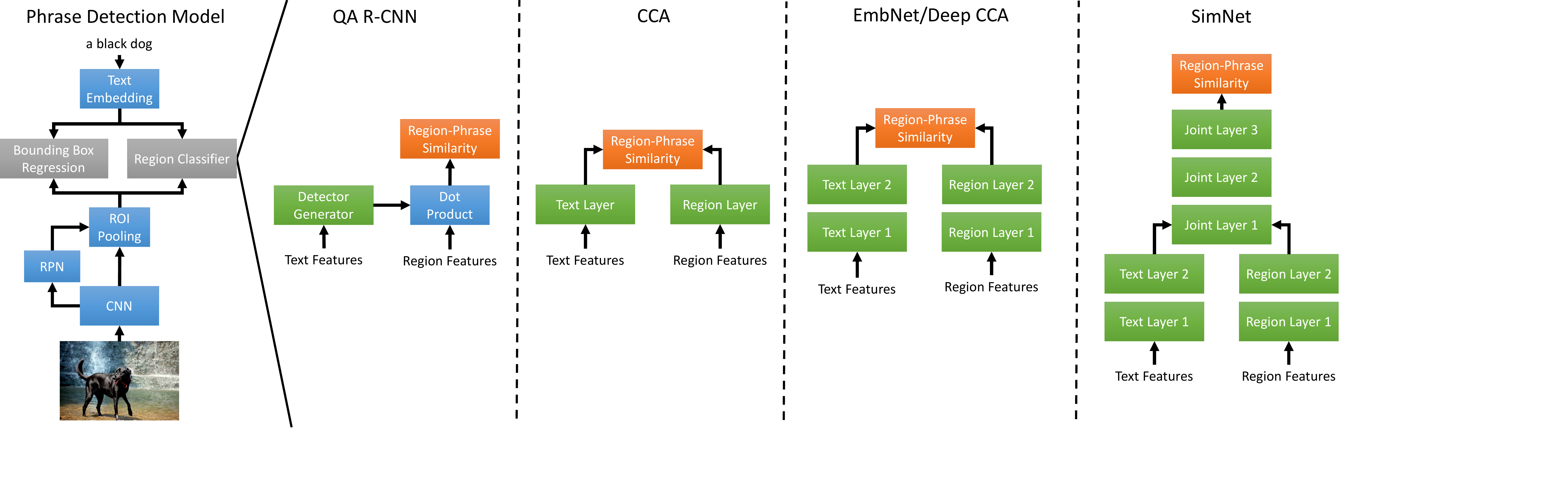}
\caption{{\bf Model Overview.} Our phrase detection model follows the Faster R-CNN paradigm (shown on the left) consisting of a region proposal network, a bounding box regressor, and a region classifier. 
For the region classifier, which separates regions that are relevant to a phrase from irrelevant regions, we benchmark several variants on phrase detection based on methods used for the phrase localization task. We find careful initialization of the region classification layers are key to enable our model to discriminate between related phrases.}
\label{fig:method}
\end{figure*}

\section{Related Work}

Most visual grounding approaches are neural networks that fuse visual and textual representations. Text features are typically obtained from pre-trained language embeddings like Word2Vec~\cite{mikolov2013efficient}, FastText~\cite{bojanowski:hal-01154523}, or BERT~\cite{devlin2018bert}, and visual features are often obtained from pre-trained convolutional neural networks (CNNs) such as VGG~\cite{simonyan2014very} or ResNet~\cite{He2015}, which are also used as a backbone in Faster R-CNN~\cite{renNIPS15fasterrcnn} adapted to visual grounding (\eg,~\cite{hinamiARXIV2017,ChenICCV2017}). Typical fusion strategies involve learning a cross-modal embedding space between image region and phrase features where distances are meaningful~\cite{flickrentitiesijcv,wang2016CVPR,wangTwoBranch2017} or a classifier on top of a fused region-phrase representation~\cite{ChenICCV2017,fukui16emnlp,wangTwoBranch2017,plummerCITE2017}.  

As explained in the Introduction, much of the prior work in phrase localization assumes each query phrase is actually present in the image. Some papers have investigated fusion strategies for image and text features (\eg~\cite{wangTwoBranch2017,rohrbach2015,fukui16emnlp,hu2015natural,plummerCITE2017}).  Others have focused on how candidate regions are selected~\cite{yehNIPS2017,ChenICCV2017} or incorporated more sophisticated linguistic cues~\cite{Luo_2017_CVPR,plummerPLCLC2017,wang2016matching,Liu_2017_ICCV}. 
Many attention models for tasks like image captioning or visual question answering tries to associate individual words with regions of an image (\eg,~\cite{fang2014captions,MisraNoisy16,Yao_2017_ICCV,Anderson_2018_CVPR,lee2018stacked}).  However, Liu~\etal~\cite{liuAAAI2017} showed these kinds of models did not localize the individual concepts as well as supervised phrase grounding methods.


More relevant to our work are phrase grounding methods that do not assume a ground truth image-phrase pair is provided at test time. Zhang~\etal~\cite{Zhang_2017_CVPR} evaluated a scenario in which each query phrase had to be localized in all its positive images (\ie, images known to contain the phrase) plus a limited number of negative (distractor) images.  
Hinami and Satoh~\cite{hinamiARXIV2017} addressed a simpler version of phrase detection covering just the most common phrases ($< 0.001$\% of the available text queries),  effectively ignoring the challenging zero- and few-shot aspects of phrase detection.  As we will show, the largest disparity in performance across methods comes from how well they handle these uncommon phrases.  Thus, including these aspects has significant ramifications for how we evaluate compared methods.   


A task related to phrase detection is {\em dense captioning}~\cite{densecap2015} generates descriptions for image regions.  However, Zhang~\etal~\cite{Zhang_2017_CVPR} showed this model performs quite poorly on the standard phrase localization task.  This likely can be attributed, in part, to the difficulty in capturing details that has been a well-documented problem for image captioning methods. It also explains the trend on the bidirectional image-sentence retrieval task where discriminative methods tend to perform better than generative models~\cite{lee2018stacked,Wehrmann_2018_CVPR}.  The metrics used to evaluate dense captioning are also quite different. For this task,  describing a region with the phrase \emph{a young man} when it should be associated with the phrase \emph{an old man} would be considered as mostly correct, whereas in our formulation it is considered completely incorrect.

Also related are {\em visual relationship detection} (VRD)~\cite{lu2016visual}, {\em scene graph generation}~\cite{Johnson2015CVPR}, and visual query detection~\cite{acharya-etal-2019-vqd}.  In VRD the task is to determine which relationships exist in an image and localize them, but is often performed over a limited set of pre-defined objects and predicates measured in Recall@[50, 100].  One need only localize a single instance of a relationship for it to be considered correct even if multiple instances are present in an image, and images with few relationships can have many incorrect predictions without penalty. Like phrase detection, scene graph generation identifies large vocabularies of concepts in images. However, like phrase localization, scene graphs are evaluated on how well they localize known entities in an image and not on their ability to discriminate between images that contain a phrase and those that do not.  In contrast, phrase detection uses an open-ended vocabulary with metrics that take into account how well a model separates all correct and incorrect predictions.  Visual query detection contains questions asking about variable numbers of objects, their color, or their positions.  However, phrase detection may require jointly reasoning about all three types of questions, as well as identify novel phrases at test time.
\section{Phrase Detection Model}
\label{sec:model}

This section describes the models we use in our study of phrase detection. As our backbone, we adopt the Faster R-CNN~\cite{renNIPS15fasterrcnn} architecture, as illustrated in Figure \ref{fig:method}. First, as in the standard Faster R-CNN, an initial image representation is computed using a convolutional neural network (CNN), and the resulting feature map is fed into a region proposal network (RPN) that generates a set of candidate bounding boxes (Section~\ref{sec:rpn}).  A single set of proposals is generated per image, \ie, the RPN is agnostic to the phrases being detected.  Next, we use an ROI Pooling layer~\cite{girshickICCV15fastrcnn} to obtain a feature representation for each region proposal selected by the RPN. In the standard Faster R-CNN, the ROI feature gets fed into a bounding box regressor (BBReg) and region classifier to predict a refinement to the bounding box coordinates and a set of per-class probabilities for the region. To extend this architecture to phrase detection, we introduce a {\em text embedding} branch that computes an encoding of the target phrase, and its output is fused with the ROIPool feature to compute a joint region-phrase representation that is used by both the BBReg and region classifier layers. In particular, the output of the region classifier is the relevance score between the phrases being detected and the image regions (Section~\ref{sec:classification}).  When obtaining scores for multiple batches of phrases, the CNN representation and RPN/ROI Pooling need only be computed once since they are not phrase-specific, significantly reducing the computational cost of testing a large number of phrases for each image. 

As discussed in the Introduction, discriminating between closely related phrases is a key challenge of phrase detection. Accordingly, the bulk of our experimental study consists in comparing a number of region classifiers adopted from phrase localization literature, as illustrated on the right of Figure \ref{fig:method}. We use the same RPN and BBReg components to generate region proposals for all classifiers we compare. Sections \ref{sec:rpn} and \ref{sec:bbreg} will describe the design of the RPN and BBReg components in detail, while Section \ref{sec:classification} will introduce all the region classifiers included in our study. Later, Section \ref{sec:enhancements} will discuss additional implementation aspects we found to be important for improving phrase detection performance, including initialization of region-phrase alignment layers, and augmentation and sampling procedures for phrases to deal with data sparsity and the long-tailed distribution of phrases in the training data.


\subsection{Region Proposal Network}
\label{sec:rpn}

Rather than using a hand-crafted category-independent region proposal method like many earlier phrase localization approaches (\eg~\cite{hu2015natural,plummerCITE2017,flickrentitiesijcv,rohrbach2015}), we train an RPN followed by a phrase-aware bounding box regression to obtain region candidates. We follow the original RPN formulation~\cite{renNIPS15fasterrcnn}, which we shall briefly review.  The RPN predicts the proposals most likely to contain objects from a set of anchor boxes.  These anchors are generated over an image feature map output by the CNN using different scales and aspect ratios.  Positive anchor boxes are those with at least 0.7 intersection over union (IOU) with a ground truth box. 

The parameters of the RPN are trained using a weighted linear combination of a log-loss over two labels indicating whether an anchor contains an object or not, along with smooth $L_1$ (\ie Huber) loss~\cite{girshickICCV15fastrcnn,huberloss}. 
 Adopting the notation from Ren~\etal~\cite{renNIPS15fasterrcnn}, let $t_i$ be the predicted box for anchor $i$, $t_i^*$ the ground truth box, $p_i$ the likelihood of $t_i$ being an object, and $p_i^*$ the indicator variable that is 1 if the anchor is positive and 0 otherwise.  
 The RPN loss is then defined as:
\begin{dmath}
L_{RPN} = \frac{1}{N_c}\sum_i^{N_c} L_{log-loss}(p_i, p_i^*) + \frac{\lambda_2}{N_r}p_i^*\sum_i^{N_r} L_{smooth L1}(t_i, t_i^*).
\end{dmath}
\noindent where $\lambda_2$ is a scalar parameter and $N_r, N_c$ are the number of samples in a minibatch and the number of anchor locations, respectively.  In our experiments, we kept all RPN-related hyperparameters the same as their defaults for training an object detector on the COCO dataset~\cite{lin2014microsoft} in a publicly available implementation of Faster R-CNN\footnote{\url{https://github.com/endernewton/tf-faster-rcnn}}.  We use a 101-layer ResNet~\cite{He2015} as our CNN image encoder and initialize it with a network that was trained for object detection on COCO~\cite{lin2014microsoft}.  During training we randomly subsample five ground truth phrases per image as we found having balanced minibatches improves performance, except on ReferIt where we found subsampling two phrases performs better.

\subsection{Phrase-Aware Bounding Box Regression}
\label{sec:bbreg}

As stated in Section \ref{sec:model}, BBReg and the region classifier take as input a joint image-text representation obtained by fusing ROIPool features representing the region proposal with an embedding of the target phrase. The BBReg component is more straightforward, so we discuss it first.


Our phrase encoding is given by HGLMM Fisher vectors~\cite{klein2014fisher}, which are built on top of word2vec~\cite{mikolov2013efficient} and PCA-reduced to 6K-D.  Then they are projected to the same size as the region features using a fully connected layer with batch normalization~\cite{Ioffe:2015:BNA:3045118.3045167}. This representation is partly a legacy of our earlier work~\cite{plummerPLCLC2017,plummerCITE2017}, but is also supported by our concurrent study~\cite{burnsLanguage2019}, in which some of the present co-authors have discovered that HGLMM features often outperform more recent embeddings on vision-language tasks. 

The input to the bounding box regressor is given by the element-wise product of the above region and phrase features. While our feature representation is different, our loss for BBReg is the same as in Ren~\etal~\cite{renNIPS15fasterrcnn}, \ie,
\begin{equation}
L_{reg} = \frac{1}{4N_r}\sum_i^{N_r} L_{smooth L1}(t_i, t_i^*).
\end{equation}
We use a single fully connected layer to for BBreg, and train it jointly with the RPN first. Later we initialize the classification layers and train them as discussed in the next section. At the end, we fine-tune the whole network. 

\subsection{Region classifier}
\label{sec:classification}

The region classifier's task is to output a confidence or compatibility score given region and phrase features. The region and phrase features are the same as those used by BBReg (Section~\ref{sec:bbreg}). For the fusion and classification of these features, we compare several methods that show the most promise in current phrase localization literature. Note that each region classifier uses the same bounding box proposal method to provide a fair comparison (the upper bound performance of these proposals is provided in Table~\ref{tab:phrase_localization}).

\subsubsection{Query Adaptive R-CNN (QA R-CNN)}
\label{subsec:qa_rcnn}

QA R-CNN~\cite{hinamiARXIV2017} is an earlier adaptation of Faster R-CNN to phrase grounding. To relate image regions to phrases, QA R-CNN generates the parameters of a linear classifier given the text features as input (see Figure \ref{fig:method} for an illustration).  More formally, let $w_c$ be the weights of a linear classifier and $v$ be the phrase features.  The classifier is generated using $w_c = W_cv$, where $W_c$ is a learned projection.  During training, each phrase in an image is considered its own category in a sigmoid cross-entropy loss. We use a single fully connected layer for BBReg instead of a multi-layer perception as in~\cite{hinamiARXIV2017} we found it produced similar results. 
\smallskip

\noindent\textbf{Negative Phrase Augmentation (NPA).} An important component of~\cite{hinamiARXIV2017} is an approach for sampling negative phrases during training to make the model more discriminative.  For every phrase associated with an image, the idea is to find the phrases a model is most often confused with and add some of them to a minibatch as ``hard negative phrases.'' In practice, these hard negative phrases are obtained from a confusion table that is updated every 10K iterations and contains 500 hard negatives for phrases in the training set.  

A big potential problem with NPA is that, since phrase grounding datasets are very sparsely labeled, many putative hard negative phrases are likely unlabeled positive examples. Hinami and Satoh~\cite{hinamiARXIV2017} proposed two ways of addressing this issue.  The first method is to use WordNet~\cite{Miller:1995:WLD:219717.219748} to identify hypernum relationships and remove phrases with a parent-child relationship (\eg, \emph{a person} couldn't be a hard negative for \emph{a man}).  However, many phrases that refer to the same object could still pass this test, \eg, \emph{a woman} could still be considered a hard negative phrase for \emph{a skier}.  This led Hinami and Satoh to propose using the dataset annotations to identify these mutually non-exclusive phrases.  If two phrases were often annotated as referring to the same object, then they also could not be used as hard negatives for each other.  While this could work for common phrases, to which~\cite{hinamiARXIV2017} restricts its evaluation, in our unrestricted phrase detection scenario, many phrases occur very few times, which means NPA would mostly rely on WordNet to identify false negative phrases.  We manually inspected the entries of 30 randomly selected phrases in the confusion table when trying to train NPA using our best phrase detection model, and found that 21 had obvious false negatives within the top few most confused phrases. 
Thus, we expect the phrase confusion table to be quite unreliable. Accordingly, in our experiments reported in Section~\ref{sec:det}, using NPA leads to negligible performance differences on phrase detection, while also increasing training time.

\subsubsection{Canonical Correlation Analysis (CCA)}
\label{subsec:cca}

CCA~\cite{hotelling1936relations} is a classical method often used to benchmark vision-language tasks (including for phrase localization~\cite{flickrentitiesijcv}).  The goal of CCA is to learn linear transformations $U, W$ between two sets of paired variables $X, Y$ (in our case, representing the image region and text features) that maximizes the correlations between them, \ie,
\begin{equation}
\begin{aligned}
    \max & \text{ tr}(W^TX^TYU)\\
    \text{subject to:} & \text{ } W^TX^TXW = U^TY^TYU = I.\\
\end{aligned}
\end{equation}
\noindent CCA can be solved as a generalized eigenvalue decomposition problem, where the eigenvectors of the top eigenvalues are concatenated to form the projection matrices.  In our experiments we use {\em normalized CCA}~\cite{gong14}, which scales the learned projection matrices by the eigenvalues, which is well known to give better performance than regular CCA. At test time, we apply the learned CCA transformations to region and phrase features, scale these projected features, and then use cosine similarity to score regions and phrases. By the very definition of CCA, only positive region-phrase pairs are used when learning the projection parameters, and typically the entire dataset is used in a single batch.  Despite CCA's simplicity, our experiments will show that it is robust to the long-tailed distribution of phrases in existing datasets, making it a strong baseline for phrase detection.

\subsubsection{Deep CCA} \label{subsec:deepcca}
Deep CCA~\cite{deepcca} is intended to address two deficiencies of traditional CCA, namely, that it is a linear projection method, and that its objective cannot be back-propagated to the feature representation. Deep CCA uses a correlation loss to train the feature representation before using CCA to learn the final transformation. At training time, a singular value decomposition is computed over features in a minibatch to form an approximation of the data covariance matrix. This requires a relatively large minibatch that must also increase with the dimensionality of the desired embedding. The resulting GPU memory requirements make it difficult to train an embedding of the same dimensionality as that of linear CCA.  Ultimately, we found that linear CCA outperforms Deep CCA even when they have the same output dimensions. More specifically, in our experiments, we needed a batch size of 30K to keep the loss numerically stable when learning a feature embedding of 1,024 dimensions. Thus, we kept the underlying CNN fixed when learning this feature embedding.  Unlike Andrew~\etal~\cite{deepcca}, which used three fully connected layers to learn their feature representation, we got the best performance with a single FC layer.


\subsubsection{Embedding Network (EmbNet)}
The Embedding Network~\cite{wangTwoBranch2017} is a fairly general way to fuse image and text features for cross-modal retrieval and classification tasks. Our implementation (refer back to Figure \ref{fig:method} for an illustration) consists of two fully connected layers each for the region and phrase features, projecting them into a shared embedding space.  The projections are trained with a triplet loss. Given some query phrase $q$, a positive region $r_p$ and negative region $r_n$, our loss is
\begin{equation}
L_T{_{qr}}(q, r_p, r_n) = \max\{0, m + d(q, r_p) - d(q, r_n)\},
\label{eq:triplet}
\nonumber
\end{equation}
\noindent where $d$ is the Euclidean distance and $m$ is a scalar parameter representing a minimum margin between positive and negative pairs.  Given some phrase $q$, a positive region during training $r_p$ is defined as a region with at least 0.6 IOU with the ground truth.  
After the model is trained, we define the confidence score for a region and a phrase as the distance between the $L_2$-normalized embedded vectors. To get the best performance, in addition to the cross-modal triplet loss above, we also include within-modality terms as in~\cite{wangTwoBranch2017}, imposed on triplets of regions and phrases, respectively.  


\subsubsection{Similarity Network (SimNet)}
\label{subsec:cite}
The Similarity Network~\cite{wangTwoBranch2017} consists of two components.  First, it learns a good feature embedding for the image regions and phrases just like EmbNet.  Then, it learns a metric to compare these features by performing an elementwise multiplication of the image region and phrase features and feeding them into three of fully connected layers, the last of which measures the similarity between the image regions and phrases (\ie, its output is a single number).  This network is trained using a logistic loss with  $L_1$ regularization on the conditional weights. Let $K$ be the number of image region-query pairs in a batch, $c$ be the confidence in a region-query pair, $l$ be its $-$1/1 label indicating whether it is a negative/positive pair, and $a$ be the conditional embedding weights before the softmax. Then our loss is:
\begin{equation}
L_{cls} = \sum^K_{i=1}\log(1 + \exp{(-l_{i}c_{i})}) + \lambda_1 L_1(a_i),
\end{equation}
\noindent where $\lambda_1$ is a scalar parameter.  We keep all hyperparameters and training procedures the same as in Plummer~\etal~\cite{plummerCITE2017}, whose implementation outperformed that of the original paper, except for our minibatch construction. Our minibatches are built on a per-image basis, so each time an image appears in a minibatch, it is also matched to all its related phrases.  Plummer~\etal sampled image-phrase pairs, so two phrases matched to the same image could appear in different minibatches.  While our minibatch construction does reduce training time significantly since each image is processed only once in each epoch, we found this hurts localization performance by about 1\%.

\section{Addressing Challenges of Open-Vocabulary Detection}
\label{sec:enhancements}

The most significant difference between phrase detection and the classical task of object detection is the need to support an open vocabulary.  In this section we address two of the associated challenges and how we address them.  Namely, in Section~\ref{sec:cca_init} we discuss our CCA initialization procedure that helps learn a more discriminative representation for the long-tailed distribution of phrases and Section~\ref{sec:pos_aug} describes how we handle label sparsity.

\subsection{Initial Vision-Language Alignment}
\label{sec:cca_init}

We obtain the best results with EmbNet and SimNet when we initialize their projection layers using all available data, instead of learning strictly from minibatches containing a fraction of the training phrases at a time. 

We assume we have two views (\ie, phrase and region features) that we would like project using a pair of fully connected (FC) layers (one for each view) so they share a single semantic space.  Network layers are initialized recursively starting with the lowest layer first which ensures the input views contain a reasonable representation either from some pre-training procedure or layers that were previously initialized using our approach.  Each layer $h(x)$ (here, $x$ can refer either to region or text features) has the following form:
\begin{equation}
    h(x) =  W(x - \mu) \sigma + b.
\end{equation}
We estimate the layer's parameters $W, \mu, \sigma$ using normalized CCA~\cite{gong14}, and initialize $b$ with zeros. The CCA objective, whose goal is to maximize the correlations between the two input views, can be solved using a generalized eigenvalue decomposition.  The eigenvectors for the top $K$ eigenvalues $\sigma$ are concatenated to form $W$, $\mu$ is the mean of the input features $x$ estimated over the entire training set, and $\sigma$ is used to scale the output features which has been shown to improve performance.  During training we only update the parameters $W, b$ and keep $\mu, \sigma$ fixed. Letting the network update $W, b$ arbitrarily, however, may result in catastrophic forgetting.  To avoid this issue, we use $L1$ regularization on $b$ (since it is zero-initialized) and $L2$ regularization between the initial CCA-estimated projection $W'$ and the updated parameters $W$, \ie,

\begin{equation}
    L_{reg} = \lambda_2\norm{W - W'}_2 + \norm{b}_1,
\end{equation}

\noindent where $\lambda_2$ is a scalar parameter that is set via grid search using validation data.  We also experimented with an iterative procedure where we alternated between estimating a FC layer's weights with CCA and fine-tuning them with the entire network, but found a single iteration was sufficient.

For EmbNet and SimNet, as illustrated in Figure \ref{fig:method} (right), the above initialization procedure is applied to Text Layer 1 and 2 and the corresponding Region Layer 1 and 2, and any other layers are randomly initialized.

\subsection{Positive Phrase Augmentation}
\label{sec:pos_aug}

Phrase grounding datasets are sparsely annotated, \ie, many regions that could conceivably be described by some phrase lack that annotation -- either because no human annotator chose to describe that particular region, or described it using a different phrase. In phrase localization this is not a critical issue, since performance is evaluated only on images already known to contain at least one instance of a phrase.  In phrase detection, however, the goal is to consistently score {\em all} instances of a phrase across the entire dataset, making it much more likely that a high-scoring negative region is simply an unlabeled positive. To help mitigate this issue, we propose a positive phrase augmentation procedure that can be used both at training time to learn a better model, and at test time to improve the accuracy of our metrics.

We begin by collecting a set of word replacements by identifying synonyms as well as more general forms of words in a phrase by extracting the hypernym relationships using WordNet.  Then, we use this set of related words as replacements for their associated words to construct additional positive labels for a phrase. For example, the phrase \emph{blue jacket} we obtain word replacements \emph{coat, cover}, and \emph{apparel} for the word \emph{jacket}.  This results in candidate positive phrases \emph{blue coat}, \emph{blue cover}, and \emph{blue apparel}.

Additional care must be taken based on how the datasets are collected. In ReferIt phrases are meant to uniquely describe an object in an image.  As a result, many phrases contain references to spatial relationships with other entities (\eg,~\emph{the box to the right of the table}).  Thus, for ReferIt we avoid replacing words like \emph{left} and \emph{right}. 
For Flickr30K Entities, however, the annotations typically refer to single entities.  This means we can consider individual words in a phrase and all combinations of them as candidates.  For example, for the phrase \emph{a large red house}, we can consider \emph{a large house}, \emph{a red house}, \emph{house}, and \emph{red} as positive examples.  On ReferIt, however, we cannot do this because breaking up a phrase like \emph{the dog on a table} would incorrectly yield both \emph{dog} and \emph{table} as positives.  After obtaining all candidate phrases associated with an image region, we only add phrases that already exist in that dataset split. 
This helps to filter out some potential false positives and odd phrase constructions.

Although NPA and PPA both use WordNet to identify words related to a phrase, PPA uses these related words to add likely positives to an image's annotations.  NPA, by contrast, uses these related words to remove false negatives. However, in practice, NPA hard negatives are often still false negatives that were missed by this procedure.

\subsection{Inverse Frequency Sampling}
\label{sec:ifs_sampling}
While every \emph{child} is a \emph{person}, not every \emph{person} is a \emph{child}, making phrases containing generic words far more common in current datasets. 
When a model sees \emph{a child} during training it can simply use features for \emph{a person}, which is seen more often. Thus, every word referring to a specific \emph{person} subgroup can use the same features, making distinguishing between them difficult.  One solution is to include hard negative phrases during training. However, as discussed in Section~\ref{subsec:qa_rcnn}, obtaining hard negatives automatically is non-trivial and prone to errors. Instead, we bias our phrase sampling procedure to prefer rare phrases that are often more specific than common phrases. In this way, we can indirectly encourage our model to learn fine-grained differences. 

Concretely, our {\em inverse frequency sampling procedure} (IFS) samples phrases with sampling budget $K$ that is the inverse of their relative likelihood in a training set.  For each image, we obtain the set of phrases and the portion of times that phrase is labeled in the dataset (\ie, if a phrase occurs 5 times out of 20 total phrase-region pairs, then it would account for 25\% of instances). Then, we renormalize all of an image's phrases so that their likelihoods sum to 1 and take their inverse.  For example, if \emph{a dog} accounts for 15\% of instances while \emph{a terrier} accounts for 5\%, and only these two phrases are associated with an image, then the likelihood we would sample \emph{a terrier} would be $1 - (0.15 / (0.05 + 0.15)) = 75$\% and \emph{a dog} 25\% of the time. We automatically select all ground truth phrases, which, by definition, are also the most specific reference to the entity (\ie, we only subsample augmented phrases from Section~\ref{sec:pos_aug}). We found a sampling budget of $K=30$ works best.
\begin{table*}[t]
\setlength{\tabcolsep}{1.5pt}
\centering
\caption{Phrase localization performance. (a) Published results of recent localization methods or reproduced using the author's code, (b) our compared approaches that use the same RPN and feature representation varying only the trained region classifier, and (c) benefits provided from initializing our classification layers using CCA.}
\label{tab:phrase_localization}
\begin{tabular}{|rl|l|c|c|c|c|c|c|c|c|c|c|c|c|}
\hline
& & & \multicolumn{4}{|c|}{Flickr30K Entities} & \multicolumn{4}{|c|}{ReferIt Game} & \multicolumn{4}{|c|}{Visual Genome}\\
\hline
& & \#Train Samples & zero-shot & few-shot & common & \multirow{2}{*}{overall} & zero-shot & few-shot & common & \multirow{2}{*}{overall} & zero-shot & few-shot & common & \multirow{2}{*}{overall}\\
& & Per Phrase & $0$ & $1-100$ & $>100$ & & $0$ & $1-100$ & $>100$ & & $0$ & $1-100$ & $>100$ &\\
\hline
\hline
\textbf{(a)} & \textbf{State-of-} & CITE~\cite{plummerCITE2017} & -- & -- & -- & 61.9 & -- & -- & -- & 34.1 & -- & -- & -- & 24.4\\
& \textbf{the-art} & G$^3$G++~\cite{Bajaj_2019_ICCV} & -- & -- & -- & 66.9 & -- & -- & -- & 44.9 & -- & -- & -- & --\\
& & FAOG~\cite{yang2019fast} & 57.2 & 56.3 & 76.1 & 67.7 & 38.9 & \textbf{68.1} & 86.3 & 58.9 & -- & -- & -- & --\\
\hline
{\bf (b)} & {\bf Region} & QA R-CNN & 60.4 & 66.6 & 76.0 & 71.1 & 39.5 & 65.5 & 78.7 & 56.6 & 34.8 & 38.3 & 32.6 & 35.8\\
& {\bf Cls} & QA R-CNN + NPA & 58.0 & 64.4 & 75.1 & 69.7& 38.4 & 64.3 & 78.2 & 55.7 & 34.1 & 37.6 & 32.0 & 35.2\\
& {\bf Method} & CCA & 50.8 & 60.3 & 72.0 & 65.7 & 28.4 & 58.6 & 78.9 & 49.4 & 27.3 & 32.3 & 28.2 & 29.5\\
& & Deep CCA & 52.5 & 58.6 & 68.7 & 63.5 & 27.9 & 57.1 & 77.6 & 48.4 & 26.2 & 31.0 & 27.8 & 28.3\\
& & EmbNet & 59.3 & 65.7 & 76.3 & 70.9 & 39.3 & 66.7 & 83.2 & 57.9 & 31.6 & 35.6 & 32.0 & 33.3\\
& & SimNet & 60.0 & 66.4 & \textbf{77.1} & 71.7 & \textbf{40.0} & 67.3 & 85.8 & \textbf{59.0} & \textbf{35.7} & \textbf{38.9} & \textbf{32.8} & \textbf{36.5}\\
\hline
{\bf (c)} & {\bf w/CCA} & EmbNet & 60.3 & 66.2 & 76.7 & 71.4 & 36.1 & 66.5 & 85.3 & 56.9 & 32.8 & 36.2 & 30.9 & 33.9\\
& {\bf Init} & SimNet & \textbf{60.5} & \textbf{67.1} & 77.0 & \textbf{71.9} & 36.5 & 67.8 & \textbf{86.5} & 57.8 & 34.5 & 37.7 & 32.1 & 35.4\\
\hline
\hline
& & Upper Bound & 96.5 & 95.8 & 97.3 & 96.7 & 87.7 & 96.7 & 99.5 & 93.3 & 80.1 & 76.1 & 62.1 & 75.4\\
\hline
& & \#Categories & 1,783 & 2,764 & 472 & 5,019 & 27,378 & 4,568 & 40 & 31,986 & 94,576 & 61,848 & 2,301 & 158,725\\
\hline
& & \#Test Instances & 1,860 & 4,373 & 8,248 & 14,481 & 29,304 & 21,850 & 14,039 & 65,193 & 100,292 & 96,600 & 41,177 & 238,069\\
\hline
\end{tabular}
\end{table*}

\section{Phrase Localization Experiments}
\label{sec:localization}

We begin by evaluating the models of Figure \ref{fig:method}, on the established task of phrase localization before moving on to phrase detection in Section \ref{sec:det}. As explained before, localization assumes we are provided a ground truth image-phrase pair, and the goal is to find the bounding box for the phrase in the image.  A localization is successful if the predicted box has at least 0.5 IOU with the ground truth.  When reporting performance, we separate phrases based on the number of training instances, breaking up the evaluation into zero-shot, few-shot ($1-100$), and common phrases ($>100$).  Only ground truth annotations are used to calculate the number of training instances (\ie without any data augmentation).
\smallskip

\noindent\textbf{Datasets.}  We use three common phrase grounding datasets in our experiments. Our first dataset is Flickr30K Entities~\cite{flickrentitiesijcv} that consists of 276K bounding boxes in 32K images for the noun phrases associated with each image's descriptive captions (5 per image) from the Flickr30K dataset~\cite{young2014image}.  We use the splits of Plummer~\etal~\cite{flickrentitiesijcv} that consist of 30K/1K/1K train/test/validation images.  
Our second dataset is ReferIt~\cite{kazemzadeh-EtAl:2014:EMNLP2014}, which consists of 20K images from the IAPR TC-12 dataset~\cite{Grubinger06theiapr} that have been augmented with 120K region descriptions.  We use the splits of Hu~\etal~\cite{hu2015natural}, which splits the train/val and testing sets evenly (\ie 10K each).  Our last dataset is Visual Genome~\cite{krishnavisualgenome}, which densely annotates 108,077 images with descriptions for image regions.  Following~\cite{Zhang_2017_CVPR,plummerCITE2017}, we split the dataset into 77K/5K/5K images for training/testing/validation, with leftover images being unused. See Table~\ref{tab:phrase_localization} for statistics on the number of phrases and instances for all datasets.  

\subsection{Localization Results}
Tables~\ref{tab:phrase_localization} reports phrase localization performance on all three datasets. Part (a) reports numbers from prior work. The last line of each subtable gives the performance of our implementation of the SimNet model using RPN and BBReg. We can see that SimNet outperforms more recent methods. Thus, we can regard our implementation of SimNet as a representative of the state-of-the-art in phrase localization.

Table~\ref{tab:phrase_localization}(b) presents a comparison of different region classifiers.  
There are only minor differences between the datasets.  For example, the zero-shot phrases tend to perform better than common phrases on Visual Genome, where the opposite was true on the other two datasets, but  overall performance across methods was relatively consistent across datasets.
We also observe that adding NPA to QA R-CNN (discussed in Section~\ref{subsec:qa_rcnn}) slightly decreases performance, which is consistent with the results in~\cite{hinamiARXIV2017}.  NPA is designed to improve discrimination of similar phrases, which is largely unnecessary in localization since similar phrases rarely occur in the same image. Notably, Deep CCA performs worse than regular CCA. As discussed in Section \ref{subsec:deepcca}, we conjecture that even a batch size of 30K is insufficient to get the CCA objective to generalize well (for comparison, for Flickr30K Entities we use 420K samples for CCA).  In addition, since Deep CCA requires a large batch size for training, fine-tuning the entire network is impractical (so all layers except those for the classifier are kept fixed). The last two lines of Table~\ref{tab:phrase_localization}(b) shows the results of EmbNet and SimNet with random initialization. We can see that SimNet outperforms EmbNet and QA R-CNN.

Table~\ref{tab:phrase_localization}(c) reports the performance of EmbNet and SimNet with CCA initialization. 
This initialization does not make much difference for localization (accuracy improves slightly on Flickr30K Entities and decreases on ReferIt and Visual Genome). However, we argue that the underlying model can better to discriminate between closely related phrases, the effect of which will be seen through consistent, larger increases in detection performance in Section \ref{sec:det}. 

Table~\ref{tab:ppa_localization} benchmarks CCA-initialized methods on new test sets created by augmenting positive phrases (discussed in Section~\ref{sec:pos_aug}), which helps reduce annotation sparsity.  The relative performance of methods remains unchanged, but the drop in absolute performance suggests the standard benchmark may overestimate a model's true performance since many of the alternative ways of referencing the same phrase result in phrases not being correctly localized, especially on Flickr30K Entities.  The last two lines of Table~\ref{tab:ppa_localization}(b) compare randomly subsampling the augmented positive phrases during training (RS) with the IFS sampling method described in Section~\ref{sec:ifs_sampling}.  Although IFS gives better performance than RS, it does result in a slight drop in performance compared with using all phrases.  However, as we discuss in Section~\ref{sec:ifs_sampling}, this is due to overfitting to phrase localization when using all phrases, making it less able to identify if a phrase is relevant to an image. This will be clearly shown when evaluating phrase detection in the next section.

\begin{table*}
\setlength{\tabcolsep}{1.5pt}
\centering
\caption{Phrase localization performance using augmented positive phrases (PPA) discussed in Section~\ref{sec:pos_aug} for evaluation.  (a) compares methods that are trained using the ground truth annotations and (b) reports the effect training with PPA has on performance.  All methods use CCA as either the region classifier or for layer initialization.}
\label{tab:ppa_localization}
\begin{tabular}{|rl|l|c|c|c|c|c|c|c|c|c|c|c|c|}
\hline
& & & \multicolumn{4}{|c|}{Flickr30K Entities} & \multicolumn{4}{|c|}{ReferIt Game} & \multicolumn{4}{|c|}{Visual Genome}\\
\hline
& & \#Train Samples & zero-shot & few-shot & common & \multirow{2}{*}{overall} & zero-shot & few-shot & common & \multirow{2}{*}{overall} & zero-shot & few-shot & common & \multirow{2}{*}{overall}\\
& & Per Phrase & $0$ & $1-100$ & $>100$ & & $0$ & $1-100$ & $>100$ & & $0$ & $1-100$ & $>100$ &\\
\hline
\hline
{\bf (a)} & {\bf w/o Train} & CCA & 41.0 & 40.0 & 57.2 & 50.0 & 23.6 & 42.5 & 70.7 & 38.2 & 26.6 & 26.6 & 25.7 & 26.4\\
& {\bf PPA} & EmbNet & 49.3 & 52.6 & 65.1 & 59.7 & 33.8 & 53.5 & 78.0 & 48.4 & 28.9 & 29.0 & 28.3 & 28.9\\
& & SimNet & 51.6 & 53.4 & 66.1 & 60.7 & 36.0 & 55.7 & 79.2 & 50.5 & 32.6 & 31.2 & 29.5 & 31.4\\
\hline
& & Upper Bound & 95.7 & 96.7 & 8.0 & 97.4 & 92.5 & 97.3 & 99.4 & 95.5 & 79.3 & 71.9 & 63.7 & 72.8\\
\hline
\hline
{\bf (b)} & {\bf w/Train} & CCA & 42.2 & 53.5 & 63.4 & 58.8 & 29.0 & 57.3 & 71.9 & 47.1 & 27.6 & 27.8 & 26.9 & 27.6\\
& {\bf PPA} & EmbNet & 50.7 & 55.9 & 67.5 & 62.4 & 39.3 & 67.6 & 81.9 & 57.4 & 32.2 & 31.9 & 29.9 & 31.7\\
& & SimNet & \textbf{52.7} & \textbf{62.8} & 70.4 & \textbf{66.8} & 41.2 & \textbf{67.7} & 81.3 & 58.2 & 34.0 & \textbf{37.4} & \textbf{31.8} & \textbf{35.0}\\
& & SimNet + RS & 51.3 & 62.3 & \textbf{70.6} & 66.7 & 37.8 & 63.2 & 79.4 & 54.5 & 33.2 & 35.9 & 31.4 & 34.2\\
& & SimNet + IFS & 50.0 & 62.0 & 69.2 & 65.7 & \textbf{41.9} & \textbf{67.7} & \textbf{82.0} & \textbf{58.6} & \textbf{34.1} & 36.7 & 31.5 & 34.9\\
\hline
& & Upper Bound & 96.5 & 96.9 & 98.2 & 97.6 & 92.5 & 97.2 & 99.4 & 95.5 & 79.1 & 71.7 & 63.4 & 72.6\\
\hline
\hline
& & \#Test Instances & 2,679 & 27,327 & 41,274 & 71,280 & 55,587 & 56,016 & 17,743 & 129,346 & 110,053 & 174,017 & 62,875 & 346,945\\
\hline
\end{tabular}
\end{table*}

\section{Phrase Detection Experiments}
\label{sec:det}

\begin{table*}
\setlength{\tabcolsep}{1.5pt}
\centering
\caption{mAP for phrase detection split by frequency of training instances. (a) our compared approaches that use the same RPN and feature representation varying only the trained region classifier (except for FAOG, which adapts the one-stage YOLO object detector~\cite{redmonCVPR2016} to phrase grounding), and (b) benefits from initializing our classification layers using CCA.}
\label{tab:phrase}
\begin{tabular}{|rl|l|c|c|c|c|c|c|c|c|c|c|c|c|}
\hline
& & & \multicolumn{4}{|c|}{Flickr30K Entities} & \multicolumn{4}{|c|}{ReferIt Game} & \multicolumn{4}{|c|}{Visual Genome}\\
\hline
& & \#Train Samples & zero-shot & few-shot & common & mean/  & zero-shot & few-shot & common & mean/ & zero-shot & few-shot & common & mean/\\
& & Per Phrase & $0$ & $1-100$ & $>100$ & total & $0$ & $1-100$ & $>100$ & total & $0$ & $1-100$ & $>100$ & total\\
\hline
\hline
\hline
{\bf (a)} & {\bf Region} & FAOG~\cite{yang2019fast} & 3.2 & 3.5 & 7.6 & 4.8 & 0.2 & 0.5 & 10.9 & 3.9 & -- & -- & -- & --\\
& {\bf Cls} & QA R-CNN & 3.9 & 4.3 & 8.9 & 5.7 & 0.2 & 0.6 & 11.8 & 4.2  & 1.6 & 1.9 & 2.6 & 2.0\\
& {\bf Method} & QA R-CNN + NPA & 3.8 & 4.1 & 9.7 & 5.9 & 0.2 & 0.6 & 11.2 & 4.0 & 1.7 & 2.0 & 2.8 & 2.1\\
&  & CCA-10 regions & 7.0 & 8.0 & 9.8 & 8.3 & 0.3 & 1.1 & 12.4 & 4.6 & 1.8 & 2.3 & 3.1 & 2.4\\
&  & CCA & 8.6 & 10.5 & 17.2 & 12.1 & 0.8 & 2.2 & 17.6 & 6.8 & 2.3 & 2.8 & 3.7 & 2.9\\
& & Deep CCA & 6.4 & 7.5 & 14.9 & 9.6 & 0.5 & 1.9 & 15.4 & 5.9 & 1.9 & 2.4 & 3.0 & 2.4\\
& & EmbNet & 3.1 & 4.0 & 9.1 & 5.4 & 0.2 & 0.6 & 12.6 & 4.5 & 1.8 & 2.1 & 2.7 & 2.2\\
& & SimNet & 4.7 & 4.4 & 8.6 & 5.9 & 0.3 & 0.7 & 13.0 & 4.7 & 1.9 & 1.9 & 2.7 & 2.2\\
\hline
{\bf (b)} & {\bf w/CCA} & EmbNet & 9.2 & 10.3 & 17.2 & 12.3 & 0.7 & \textbf{2.5} & 18.0 & 7.0 & 2.4 & 2.9 & \textbf{3.6} & 3.0\\
& {\bf Init} & SimNet & \textbf{9.7} & \textbf{11.2} & \textbf{17.3} & \textbf{12.7} & \textbf{1.1} & \textbf{2.5} & \textbf{18.5} & \textbf{7.4} & \textbf{2.7} & \textbf{3.0} & \textbf{3.6} & \textbf{3.1}\\
\hline
\end{tabular}
\end{table*}

\begin{table*}
\setlength{\tabcolsep}{1.5pt}
\centering
\caption{mAP for phrase detection split by frequency of training instances where augmented positive phrases (PPA) discussed in Section~\ref{sec:pos_aug} are used for evaluation.  (a) compares methods that are trained only using the ground truth annotations and (b) reports the effect also training with PPA has on performance.  All methods use CCA as either the region classifier or for layer initialization.}
\label{tab:ppa_detection}
\begin{tabular}{|rl|l|c|c|c|c|c|c|c|c|c|c|c|c|}
\hline
&& & \multicolumn{4}{|c|}{Flickr30K Entities} & \multicolumn{4}{|c|}{ReferIt Game} & \multicolumn{4}{|c|}{Visual Genome}\\
\hline
&& \#Train Samples & zero-shot & few-shot & common & mean/ & zero-shot & few-shot & common & mean/ & zero-shot & few-shot & common & mean/\\
&& Per Phrase & $0$ & $1-100$ & $>100$ & total & $0$ & $1-100$ & $>100$ & total & $0$ & $1-100$ & $>100$ & total\\
\hline
\hline
\textbf{(a)} & \textbf{w/o Train} & CCA & 8.9 & 10.7 & 18.9 & 12.9 & 0.8 & 2.1 & 16.0 & 6.3 & 2.3 & 2.8 & 3.6 & 2.9\\
& \textbf{PPA} & EmbNet & 8.7 & 10.4 & 19.6 & 12.9 & 1.0 & 2.0 & 16.5 & 6.5 & 2.5 & 2.8 & 3.6 & 3.0\\
&& SimNet & 9.4 & 11.2 & 19.8 & 13.5 & 1.2 & 2.3 & 17.3 & 6.9 & 2.7 & 2.9 & 3.8 & 3.1\\
\hline
\textbf{(b)} & \textbf{w/Train} & CCA & 8.4 & 10.8 & 18.8 & 12.7 & 0.8 & 2.5 & 15.7 & 6.4 & 2.2 & 2.8 & 3.6 & 2.9\\
& \textbf{PPA} & EmbNet & 8.3 & 10.3 & 19.8 & 12.8 & 0.9 & 2.5 & 16.3 & 6.6 & 2.5 & 2.8 & 3.8 & 3.0\\
&& SimNet & 8.6 & 11.4 & 21.1 & 13.7 & 1.2 & 2.8 & 17.9 & 7.3 & 2.6 & 2.9 & 3.8 & 3.1\\
&& SimNet + RS & 9.0 & 11.2 & 20.4 & 13.6 & 1.1 & 2.8 & 17.9 & 7.3 & 2.6 & 2.9 & 3.7 & 3.1\\
&& SimNet + IFS & \textbf{9.4} & \textbf{11.8} & \textbf{21.4} & \textbf{14.2} & \textbf{1.3} & \textbf{2.9} & \textbf{18.5} & \textbf{7.6} & \textbf{2.7} & \textbf{3.1} & \textbf{4.0} & \textbf{3.3}\\
\hline
\end{tabular}
\end{table*}

\begin{figure*}
\centering
\topinset{\bfseries(a) No CCA Initialization}{\topinset{\bfseries(b) w/CCA Initialization}{\includegraphics[width=0.95\textwidth,trim=0cm 2.8cm 2.4cm 0cm,clip]{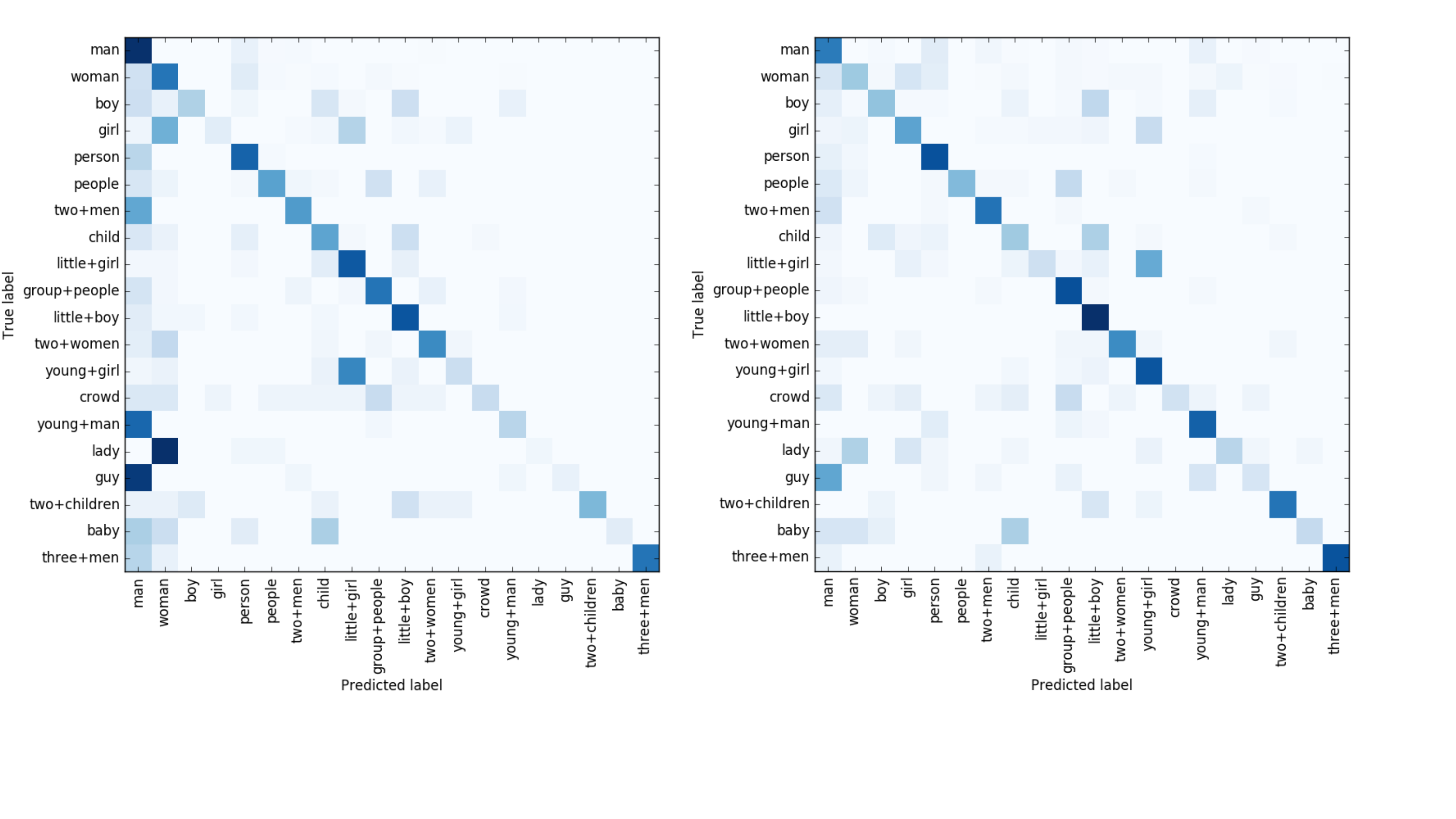}}{-.0in}{0.8in}}{-0.0in}{-2.7in}
\caption{Confusion matrix comparing the top 20 most common phrases referring to people  (in order of number of instances) in the Flickr30K Entities test set for different versions of the EmbNet classifier.}
\label{fig:confusion_matrix}
\end{figure*}

Phrase detection is akin to object detection where the phrases are categories.  Thus, use mean average precision (mAP) over the phrases for evaluation. We found keeping a single candidate per phrase per image performed the best in our experiments -- \ie, every image predicts a single location for each phrase (we provide a comparison in Table~\ref{tab:phrase} where we considered ten regions per phrase using the CCA classifier for reference).  As with the localization experiments, we split phrases into zero-shot, few-shot, and common sets.  An overall score is obtained by averaging these three mAP scores. In practice this gives higher emphasis to the common phrases that typically account for the majority of instances, but have few unique phrases, when evaluating a model.

\subsection{Detection Results}
 Table~\ref{tab:phrase} reports performance on phrase detection on all three datasets. When comparing the different classification methods in Table~\ref{tab:phrase}(a), we see that cross-correlation methods -- \ie, CCA and Deep CCA -- significantly outperform other approaches.  This stands in direct contrast with the phrase localization experiments of Section~\ref{sec:localization}, where they performed the worse, and provides additional evidence of prior work overfitting to phrase localization.  However, as can be seen in Table~\ref{tab:phrase}(b), by fine-tuning the learned CCA weights, we can not only improve performance on phrase detection further, but, as we saw in Section~\ref{sec:localization}, train a model which is competitive with the state-of-the-art in phrase localization. As we discussed in Section~\ref{subsec:qa_rcnn}, and verified in Table~\ref{tab:phrase}(a), using NPA mostly affords benefits to common phrases .  Thus, using NPA to obtain (noisy) hard negatives during training is not as effective as CCA for improving discrimination.
 

To visualize the advantage provided by CCA, Figure~\ref{fig:confusion_matrix} compares the confusion matrices for the top 20 person phrases for the EmbNet classifier without and with CCA initialization. This verifies pure minibatch training without CCA initialization leads to a network that makes similar predictions for similar phrases, while the CCA-initialized network has much better fine-grained discrimination ability.

 Table~\ref{tab:ppa_detection} reports the performance of CCA-initialized classifiers with PPA from Section~\ref{sec:pos_aug} to reduce annotation sparsity.  We show that the relative performance of methods remains largely unchanged when using the training strategies from Table~\ref{tab:phrase}.  However, in the last line of Table~\ref{tab:ppa_detection}(b) we see that training with inverse frequency sampling (IFS) to bias to selecting harder phrases during training, as described in Section~\ref{sec:ifs_sampling}, yields a consistent improvement over using all phrases, or randomly subsampling these phrases.     

\begin{figure*}
\centering
\includegraphics[width=0.98\textwidth,trim=0cm 2.4cm 8.6cm 0cm,clip]{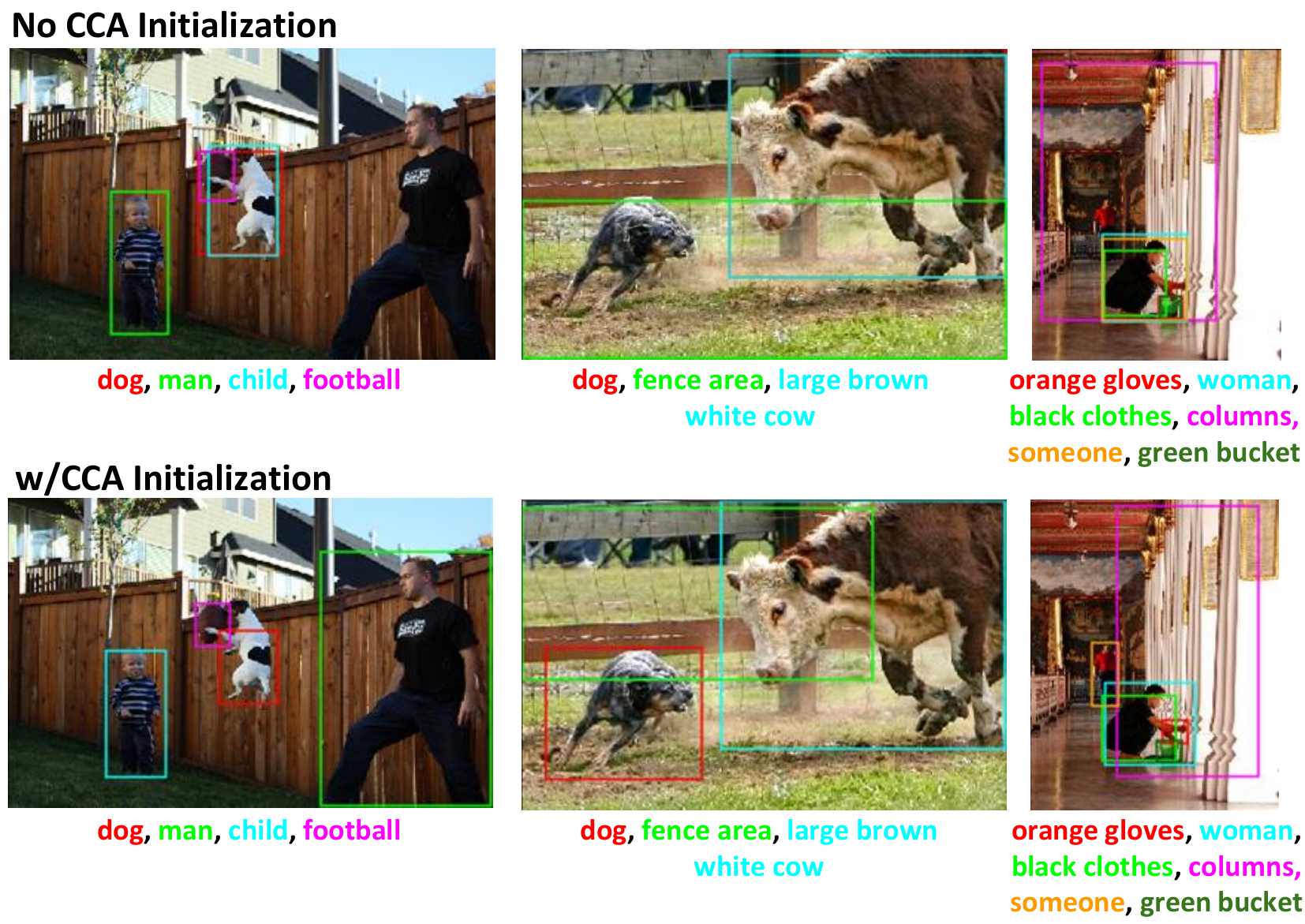}
\caption{Qualitative results comparing an Embedding Network classifier with and without CCA initialization.  See text for discussion.}
\label{fig:qualitative}
\end{figure*}

Finally, Figure~\ref{fig:qualitative} shows a qualitative comparison of the models without and with CCA initialization. In the leftmost example, before using CCA for layer initialization our model made similar predictions for the \emph{dog} and \emph{child}, while also confusing the \emph{child} and the \emph{man}, all of which we correctly identify using CCA initialized layers.  Being able to correctly identify similar phrases is not restricted to references of ``people," however, as seen in the middle example of Figure~\ref{fig:qualitative}, where the model before CCA initialization makes the same prediction for the \emph{dog} and \emph{large brown white cow}, but gets them correct with our full model.  The third example of Figure~\ref{fig:qualitative} also makes several correct references with the CCA initialized model, including for the phrase \emph{someone} even though that person is much less prominent than the \emph{woman}.  This suggests there may be some visual cues that may be useful in resolving pronominal references, and taking into account the bias of what and how entities are referenced may improve performance (\eg, human biases when writing the phrases as done in Misra~\etal~\cite{MisraNoisy16}).


\subsection{Filtering Phrases}
\label{sec:filter_phrases}

At test time, evaluating our phrase detection model for the entire phrase vocabulary may be too computationally expensive, and is likely to result in many false positive detections. To mitigate these issues, we can consider a filtering step in which we first use a global image representation to predict a short list of phrases likely to be in the image, and then selectively run our phrase detection model only on those phrases. To this end, we use the two-branch image-sentence retrieval approach of Wang~\etal~\cite{wangTwoBranch2017} trained on Flickr30K to retrieve the top 100 training sentences for each image in the test set.  Then, for each image we extract phrases from the retrieved sentences and only run the detector models for the extracted phrases.  For the text representation we use the same HGLMM features as our phrase detectors. For the whole-image representation, we use a 152-layer ResNet pretrained on ImageNet~\cite{deng2009imagenet} and averaged over 10 crops.


Table~\ref{tab:filtered_phrases} reports a consistent improvement from the above filtering procedure.  
However, a drawback of this approach is that it requires database sentences from a similar distribution of images.  We also tried to generate captions using the Show and Tell~\cite{xu2015show} approach rather than retrieve them, but we found the generated captions provided low recall on the phrases in the test set, resulting in poor performance.  

Retrieving sentences provides at least two constraints the phrase detection models lack.  First, sentences capture some information about co-occurrences between phrases (\eg, \emph{hands} often only appear when you can also see \emph{a person}).  Secondly, these sentences give some measure of the prior probability of a phrase, \ie, we are unlikely to retrieve a phrase if it occurs once in the dataset unless we are relatively certain it exists. Incorporating such constraints in an end-to-end-trainable phrase detection framework is a good potential direction for future work on phrase detection.

\begin{table}
\setlength{\tabcolsep}{5pt}
\centering
\caption{Effect the phrase filtering approach discussed in Section~\ref{sec:filter_phrases} has on the phrase detection task.  Methods are evaluated on the Flickr30K Entities test set and include PPA for both training/testing. }
\label{tab:filtered_phrases}
\begin{tabular}{|l|c|c|c|c|}
\hline
\#Train Occurrences   & zero-shot & few-shot & common & mean/ \\
Per Phrase & $0$ & $1-100$ & $>100$ & total\\
\hline
\hline
{\bf w/o Phrase Filtering} & & & &\\
EmbNet & 8.3 & 10.3 & 19.8 & 12.8\\
SimNet (w/o CCA) & 3.8 & 4.6 & 11.8 & 6.7\\
SimNet & 8.6 & 11.4 & 21.1 & 13.7\\
SimNet + IFS & 9.4 & 11.8 & 21.4 & 14.2\\
\hline
{\bf w/Phrase Filtering} &  & & &\\
EmbNet & 9.4 & 11.1 & 20.5 & 13.7\\
SimNet (w/o CCA) & 6.2 & 7.5 & 16.8 & 10.2\\
SimNet & 10.5 & \textbf{12.9} & 20.7 & 14.7\\
SimNet + IFS & \textbf{11.2} & 12.7 & \textbf{22.3} & \textbf{15.4}\\
\hline
\end{tabular}
\end{table}
\section{Conclusion}
We introduced the phrase detection task, which is more challenging and has a broader set of applications than the localization-only problem addressed in prior work.  Our experiments show that state-of-the-art localization models tend to have difficulty inferring the presence of phrases in an image compared to seemingly simpler methods like CCA. Nevertheless, by fine-tuning a CCA-initialized model with negative samples we obtain the best results on phrase detection, while also being competitive with the state-of-the-art on phrase localization.  However, our models still perform relatively poorly compared to models for tasks like object detection, indicating substantial room for improvement in future work.  A significant challenge  of phrase detection stems from the long tail of phrases that occur only a few times. As discussed in Section~\ref{sec:filter_phrases}, improvement could come from jointly predicting multiple phrases at a time while also taking into account how common a phrase is.  We believe improving negative sampling methods could have a significant impact on performance in future work. 

\ifCLASSOPTIONcompsoc
  \section*{Acknowledgments}
\else
  \section*{Acknowledgment}
\fi

This work is supported in part by DARPA and NSF awards IIS-1724237, CNS-1629700, CCF-1723379,  IIS-1718221, IIS-1563727, an Amazon Research Award and AWS ML Research Award, and Google Faculty Award. The authors would like to thank Karen Livescu for helpful discussions.

\ifCLASSOPTIONcaptionsoff
  \newpage
\fi

\bibliographystyle{IEEEtran}
\bibliography{egbib}


\end{document}